\documentclass[letterpaper, 10 pt, conference]{ieeeconf}  % Comment this line out if you need a4paper

\IEEEoverridecommandlockouts                              % This command is only needed if 
\usepackage{graphicx} % for pdf, bitmapped graphics files
\usepackage{bm}
\usepackage{amsmath}
\usepackage{amssymb}
\usepackage{amsfonts}
\usepackage{textcomp}
\usepackage{algorithm}
\usepackage{algpseudocode}
\usepackage{orcidlink}
\usepackage{array}
\usepackage{multirow}

\newcolumntype{C}[1]{>{\centering\arraybackslash}m{#1}}
\usepackage{array}
\usepackage{multirow}
\usepackage{float}
\usepackage{graphicx}
\usepackage{booktabs}
\usepackage{comment}
\usepackage{hyperref}
\hypersetup{hidelinks}

\newcommand{\blind}[1]{[ANONYMIZED]}

\newcolumntype{L}[1]{>{\raggedright\arraybackslash}m{#1}}

\title{\LARGE \bf
Active Perception of Upper-Limb Functional Mobility through Human-Robot Handovers
}

\author{
Francesco Semeraro$^{1}$\,\scalebox{1.5}{\orcidlink{0000-0002-8812-0968}},
\textit{Member, IEEE},
Marta Lagomarsino$^{1}$\,\scalebox{1.5}{\orcidlink{0000-0001-9121-1812}},
\textit{Member, IEEE},\\
Alberto Ranavolo$^{2}$\,\scalebox{1.5}{\orcidlink{0000-0002-0197-6166}}
and
Arash Ajoudani$^{1}$\,\scalebox{1.5}{\orcidlink{0000-0002-1261-737X}},
\textit{Senior Member, IEEE}
\thanks{Corresponding author: Francesco Semeraro (Email: {\tt\small francesco.semeraro@iit.it}). \newline This work was supported in part by the Italian National Institute for Insurance against Accidents at Work (INAIL) ergoCub CORE project, and by the Italian Ministry of University and Research (MUR) under the Fondo Italiano per la Scienza (FIS), call FIS 3, project EPIC, with code FIS-2024-02654.}% <-this % stops a space
\thanks{$^{1}$Francesco Semeraro, Marta Lagomarsino and Arash Ajoudani are with the Human-Robot Interfaces and Interaction (HRI$^{2}$) Laboratory, Italian Institute of Technology, 16163, Genoa, Italy.}%
\thanks{$^{2}$Alberto Ranavolo is with the Department of Occupational and Environmental
Medicine (DiMEILA), Epidemiology and Hygiene, INAIL, 00078, Rome, Italy. }%
}

\begin{document}

\maketitle
\thispagestyle{empty}
\pagestyle{empty}

%%%%%%%%%%%%%%%%%%%%%%%%%%%%%%%%%%%%%%%%%%%%%%%%%%%%%%%%%%%%%%%%%%%%%%%%%%%%%%%%
\begin{abstract}

The reintegration of people in the workplace after an upper-limb injury is often a long process. Collaborative robots and humanoids operating in shared workspaces must adapt their behaviour to the physical capabilities of their human partners. Achieving this requires robots to autonomously estimate task-relevant human mobility during interaction, without relying on external clinical assessments. In this work, we formulate the estimation of upper-limb functional mobility as an active perception problem. We model task-available upper-limb functional ranges of motion through impairment coefficients that parameterise reductions of healthy joint limits, and integrate this representation into a human model used by the robot during interaction. Based on this formulation, we propose an online active perception pipeline in which the robot actively probes the user through repeated handover requests while progressively updating a Bayesian belief over the impairment coefficients. This enables the robot to continuously refine its internal model of the user's joint-level capabilities through interaction rather than passive observation. The proposed framework was evaluated through simulation studies and a proof-of-concept physical validation with a robotic platform and enacted human impairments, demonstrating that the pipeline can accurately recover the impairment coefficients during real human-robot handovers.

\end{abstract}

%%%%%%%%%%%%%%%%%%%%%%%%%%%%%%%%%%%%%%%%%%%%%%%%%%%%%%%%%%%%%%%%%%%%%%%%%%%%%%%%
\section{INTRODUCTION}
\label{1::intro}
Going back to work after having recovered from an upper-limb injury is a delicate moment for workers, because they might have not completely retrieved their full mobility prior to the accident. Collaborative robots and humanoids are progressively appearing in industrial environments and could facilitate this process by adapting their behaviour to the physical capabilities of individual workers \cite{vianello2021human, lagomarsino2025intuitive}. %expected to physically interact with human workers at close proximity \cite{vianello2021human}. 
Such adaptations require robots to estimate the user's residual functional mobility. 
With such knowledge, robots can further tune the physical interaction with the user to make their reintegration in the workplace more manageable and personalised. Additionally, robots have the potential to plan their own motions and collaborate\begin{figure}[thpb]
      \centering
      \includegraphics[scale=0.21]{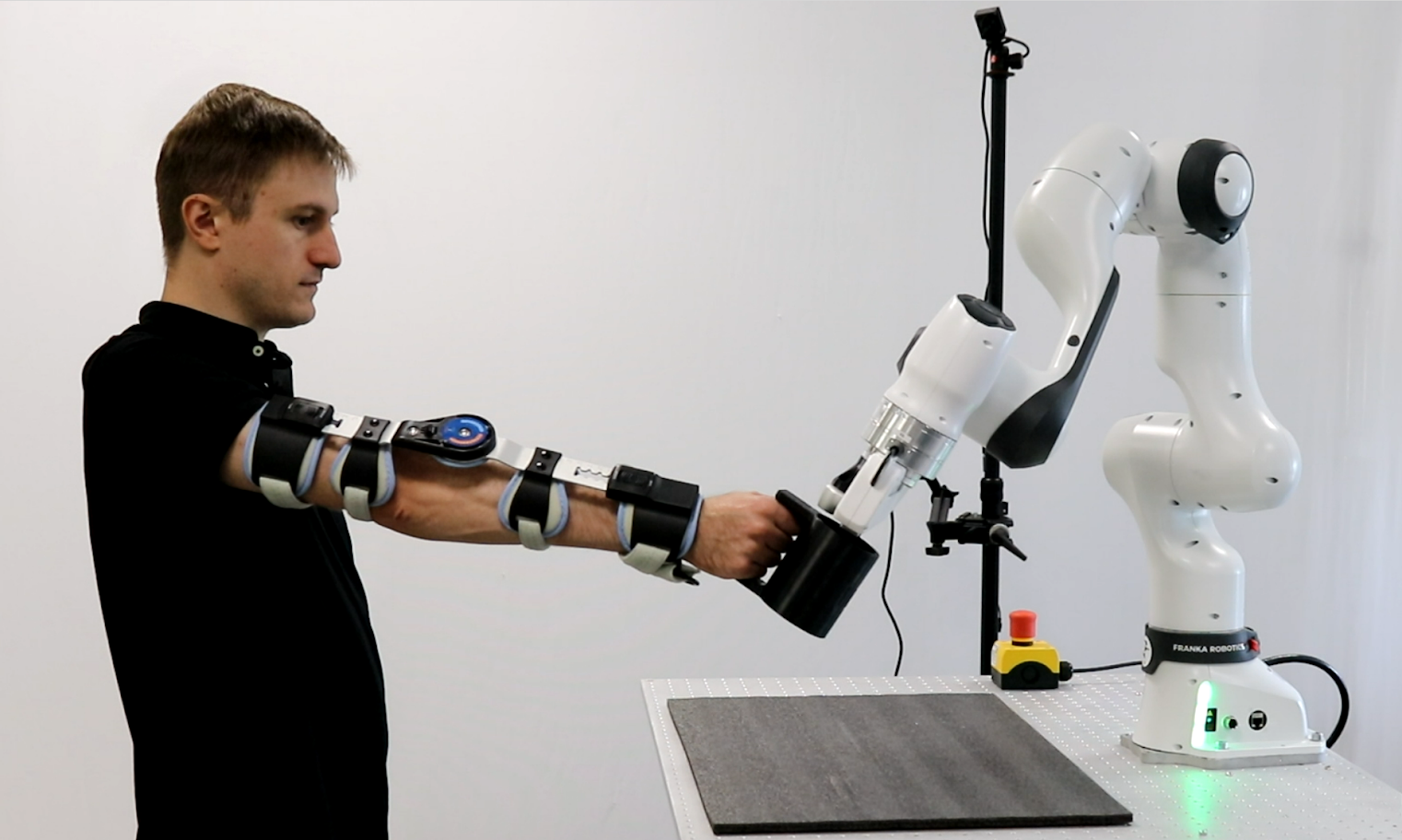}
      \caption{A person with an impairment enacted by means of a harness performs a handover requested by the robot. An underlying optimisation process computes the inverse kinematics of the human arm to the best extent allowed by the current belief on the impairment. This last gets updated based on the mismatch between the estimate and the real upper-limb joint values measured during the interaction.}
      
      %\caption{A simulated human model is set with an impairment configuration of the joints based on its current belief. Within the action space (bigger cube), the model is requested to reach a pose, represented by the smaller cube with a frame. The underlying optimisation computes the inverse kinematics of the model to the best extent allowed by the current belief on the impairment. This is updated based on the mismatch between the real joint measurements and the most likely ones based on the current belief.}
      \label{figurelabel}
   \end{figure} with workers to avoid compensatory movements and take over the workers' more proper motions related to work activity, because their functional mobility cannot yet be fully reengaged \cite{rocha2025post,lagomarsino2025mitigating}. Although robotic solutions that can account for compensatory movements have been developed in rehabilitation settings \cite{cai2024compensation}, a system capable of achieving the same purpose in industrial settings has yet to be designed. Indeed, the few existing works addressing upper-limb impairments in human-robot collaboration primarily substitute the user's movements with robotic assistance or minimise the need for arm motion altogether \cite{torielli2024laser, poirier2019voice, ardon2021affordance}, rather than explicitly modelling and adapting to the user's joint-level limitations.
%Going back to work after having recovered  an injury is a delicate moment for workers. They have not completely recovered the whole mobility prior to the accident, though this is somewhat recoverable. Human-robot collaboration can help towards a timely full recovery of the mobility. Through an active recuperation strategy, they can encourage the worker to engage the full range of motion (ROM) of their joints during the collaboration in the workplace, to induce recovery of their full mobility. By doing so, a recovery potentially started in a clinical setting with a physiotherapist can be transferred and completed in an industrial one, with the robot in charge of maintaining the recovery process. The continuation of the rehabilitation process in the workplace avoids compensatory movements to take over workers' proper motions related to work activity, due to their unwillingness to continue the recovery process on their own \cite{rocha2025post,lagomarsino2025mitigating}. While robotic solutions that can account for compensatory movements were developed in rehabilitation settings \cite{cai2024compensation}, a system capable of achieving the same purpose in industrial settings is yet to be designed.

In order to adapt an interaction to a worker's physical capabilities, the robot requires a quantitative representation of the user's mobility at the joint level. A joint's range of motion (ROM) describes its overall admissible angular interval, whereas physical impairments may restrict the portion of this interval that a user can effectively engage. In this work, we refer to this interval constrained by the impairment as the functional range of motion (fROM), which represents an estimate of the user's available joint mobility rather than the angular range necessarily expressed during a specific task. In clinical settings, standardized assessments such as the Fugl-Meyer Assessment~\cite{see2013standardized,gladstone2002fugl} evaluate sensorimotor impairment through ordinal scores based primarily on observed performance. Although these assessments can be administered without instrumented motion tracking, their scores do not directly provide the quantitative joint-level mobility bounds \cite{Chini2022}, required for human-aware motion planning. With this quantitative information, a robot could properly tune the interaction according to the specific joint movements to elicit from the user \cite{lagomarsino2026towards}. Furthermore, clinical scores are measured while having patients performing unnatural movements, which could alter the fROMs during more usual work activity. In relation to the upper limb, previous attempts of measuring its fROMs have already been performed in literature. Keyvanian et al. \cite{Keyvanian2023} explored an unsupervised learning approach to learn ROM boundary functions by collecting full-range movements of healthy people, exploring one by one all the possible ROMs of the upper arm. They then defined an impairment index, a high-level quantitative metrics of the user's impairment that takes into account all the fROMs estimated altogether. However, in this approach, they explored free movements of the arm without focusing on goal-oriented and manipulation tasks \cite{lauman2025assessing}, having participants extensively move each joint at a time to their fullest extent. Considering the upper limb, the engagement of each of its degrees of freedom (DOFs) changes from using them one at a time and altogether within a goal-oriented action \cite{Khanna2023}. Therefore, the fROMs of the user acting in a setting may differ. Because of this, Liu et al. \cite{Liu2025} proposed a framework that leveraged information from task-based recordings of people with different fROMs. Upon receiving clinical assessment scores of the ROMs related to the recordings \cite{gladstone2002fugl}, they learned a mapping from the clinical scores to the fROMs, so that the robot could learn the fROMs of a new user by using their clinical scores and without motion tracking. Although effective, this methodology relies on having the clinical scores available and updated. While these could be available at the beginning of the reintegration in the workplace, it can be demanding to summon a clinician again in the future to track and monitor the progress of the recovery outside of the clinical settings. A robot should be able to evaluate the ROMs of the user independently, without having to rely on external human inputs for this information. This dependence on external assessments may also limit how frequently the user's evolving functional capabilities can be reassessed during their return to work.

%Therefore, in this work, we built a pipeline for a collaborative robot to estimate the ROMs of the upper arm of a person through an active perception process (see Fig. \ref{figurelabel}). In this, the robotic system repetitively hands over an object to the human user and, based on the user's kinematics at the handover point, the estimate of the ROMs is updated. This process is repeated until the system finds an accurate estimate of the ROMs inspected. No previous learning is requested to perform this estimation. We demonstrated the validity of this approach through an experimental validation in which a participant performed the requested handovers while having their upper arm movements artificially hindered. The contributions of this work are threefold:
Therefore, in this work, we developed a pipeline through which a collaborative robot estimates the impairment coefficients associated with the user's upper-limb fROM through an active perception process (see Fig. \ref{figurelabel}). In this, the robotic system repetitively hands over an object to the human user. Based on the user's kinematics at each handover pose, the belief over the impairment coefficients gets updated. The resulting coefficients parameterise the modelled fROM bounds of the considered joints. This process is repeated until the system finds an accurate estimate of the fROMs inspected. No previous learning is requested to perform this estimation. We demonstrated the validity of this approach through an experimental validation in which a participant performed the requested handovers while having their upper-limb movements artificially hindered. The contributions of this work are threefold:
% \begin{itemize}
%     \item a representation of impairment coefficients that links
%     task-related upper-limb kinematics to functional joint limits;
%     \item an active probing pipeline that updates
%     these coefficients online through human-robot handovers;
%     \item a validation comprising a systematic simulation study and a
%     proof-of-concept physical experiment with enacted impairments.
% \end{itemize}

\begin{itemize}
    \item a robot-centric formulation of functional upper-limb mobility through impairment coefficients;
    \item an online active perception pipeline where the robot initiates handovers and updates a belief over the user's functional mobility;
    \item validation of the proposed framework through deployment on a real robotic platform interacting with a participant with enacted impairments.
\end{itemize}

This paper is structured as follows. Upon having introduced our work (see Section \ref{1::intro}), Section \ref{2::mm} describes the human modelling adopted for the problem and the active perception pipeline used for the fROM perception problem. Section \ref{3::exp} and \ref{4::results} report the experiments carried out and the related results, respectively. Finally, Section \ref{5::discussion} reasons about the interpretations of the results and Section \ref{6::conclusions} summarises the content of the paper.

\section{MATERIALS AND METHODS}
\label{2::mm}

\begin{figure}[!t]
      \vspace{5mm}
      \centering
      \includegraphics[scale=0.4]{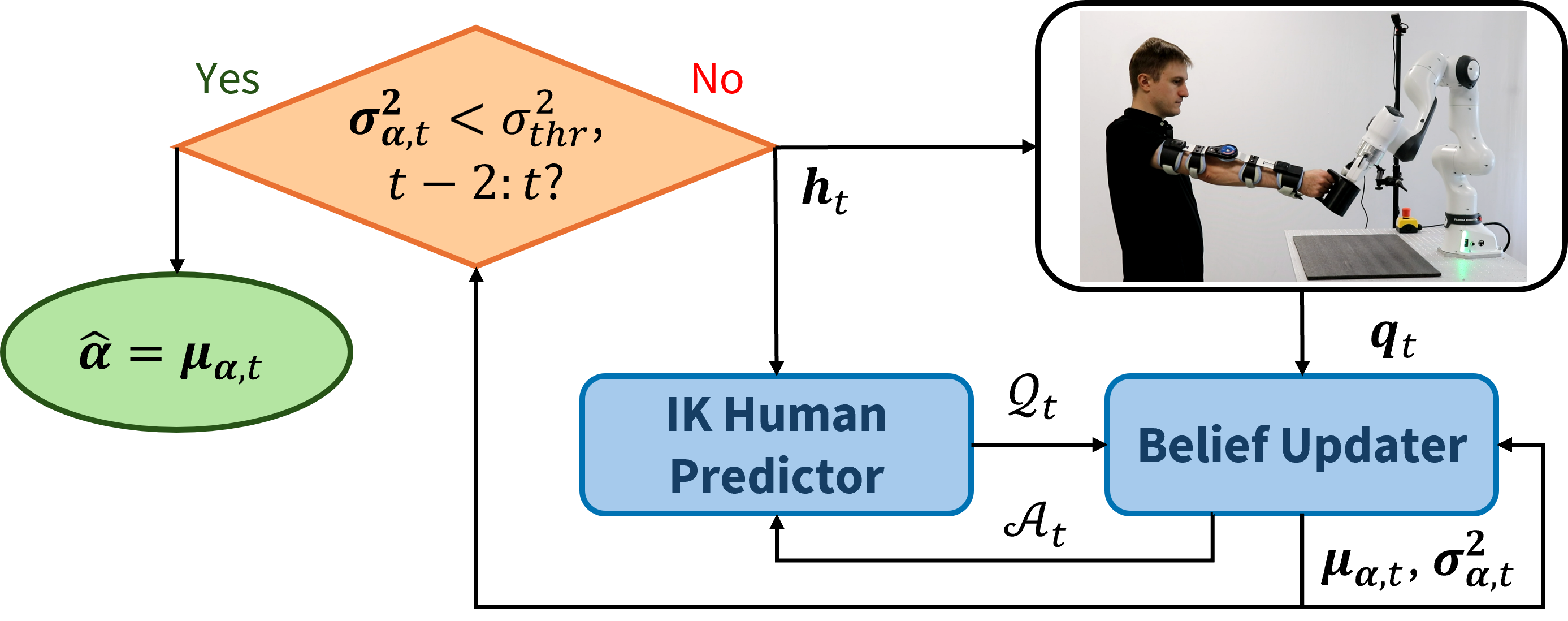}
      \caption{Diagram of the active perception pipeline for fROM estimation.}
      \label{pipeline}
   \end{figure}

The fROM active perception process was structured as follows. The robot moves the object to hand over to a designated pose in space and holds it in place, waiting for the user to reach out for it. While doing so, the robot predicts the user's kinematics at the handover point based on the object's target pose and its current belief about the user's fROMs of interest. When their hand reaches the closest possible to the handover point, the kinematics of the user gets recorded. This recording and the robot's estimate of the human kinematics are used to update the current belief on the fROMs of interest. This process is repeated until a condition on the estimate is met (see Subsection \ref{2::mm::active-perception}) and the perception process terminates. The pipeline is reported in Fig. \ref{pipeline} and its components are described below.

\subsection{Problem definition}
\label{2::mm::problem}
In our design, the user's ROMs are modelled as in \cite{lagomarsino2026towards}:
\begin{equation}
\begin{aligned}
\bm{q}
&\in
\left[
\bm{q}_{\min} + \Delta \bm{q}_{\min},
\;
\bm{q}_{\max} - \Delta \bm{q}_{\max}
\right],
\label{eq1}
\end{aligned}
\end{equation}
where $\bm{q}\in\mathbb{R}^{N_j}$ contains the joint values of the human upper-limb kinematic chain, comprising $N_j$ degrees of freedom (DOFs), and $\bm{q}_{\min}$ and $\bm{q}_{\max}$ denote the lower and upper healthy ROM limits, respectively, obtained from anthropometric references~\cite{pheasant2018bodyspace}. The adjustments to these limits are defined as follows:

\begin{equation}
\begin{aligned}
\Delta \bm{q}_{\min}
&=
\bm{\alpha}
\left(
\min\left(\bm{q}_{n}-\zeta,\bm{q}_{m}\right)
-
\bm{q}_{\min}
\right),
\\[2mm]
\Delta \bm{q}_{\max}
&=
\bm{\alpha}
\left(
\bm{q}_{\max}
-
\max\left(\bm{q}_{n}+\zeta,\bm{q}_{m}\right)
\right),
\end{aligned}
\label{alpha-eq}
\end{equation}

where $\bm{q}_n$ denotes the nominal joint configuration and $\bm{q}_m$ contains the joint values measured from the user in a reference configuration, ideally close to $\bm{q}_n$. The nominal configuration was defined as a reference handover posture, with the upper arm parallel to the ground~\cite{bestick2018learning}. The $\min$ and $\max$ operators are applied element-wise and allow the reference bounds to accommodate deviations of the measured joint configuration from the nominal configuration. The parameter $\zeta$ introduces a small margin around the nominal configuration, allowing limited joint movements even under the maximum modelled restriction. In our implementation, $\zeta$ was set to 0.08727 rad, corresponding to $5^\circ$. 

The vector $\bm{\alpha}\in[0,1]^{N_j}$ contains the impairment coefficients (ICs), with 0 representing unrestricted mobility and 1 the maximum restriction permitted by the model. Each coefficient scales the displacement of both healthy ROM boundaries towards their corresponding reference bounds, applying the same relative restriction while allowing different absolute reductions. The resulting bounds in (\ref{eq1}) define the modelled fROMs under the impairment configuration $\bm{\alpha}$, rather than the joint range directly observed during a specific task.

Our active perception process aims to estimate the components of $\bm{\alpha}$ for the DOFs of interest, yielding an estimate vector $\hat{\bm{\alpha}}$ of size $N_{imp}\leq N_j$. In this work, we focused on shoulder and elbow flexion because of their prominent contribution to forward reaching \cite{mehler2017minimizing}, the movement elicited by the proposed handover task. Accordingly, the proposed pipeline does not directly observe the boundaries of each joint ROM. Instead, it estimates the ICs that parameterise the functional joint limits represented by the human model.

\subsection{Human model}
\label{2::mm::human-model}

\begin{figure}[t]
      %\centering
      \hspace{4mm}
      %\vspace{-10}
      \includegraphics[scale=0.3]{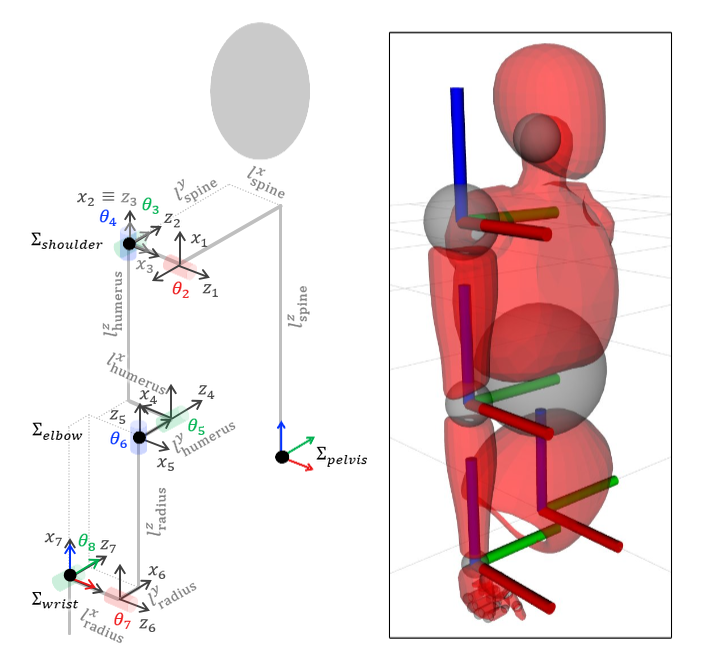}
      \caption{Kinematic chain and rendering of the human model.}
      \label{kinematic_chain}
\end{figure}

The human model considered in this work is reported in Fig. \ref{kinematic_chain} %, alongside with its Denavit-Hartenberg parameters 
\cite{lagomarsino2026towards}. When used for estimating the fROM of a specific user, the body segment sizes can be scaled accordingly. The model incorporates one IC for each considered DOF. These coefficients reduce the corresponding healthy ROM bounds according to (\ref{alpha-eq}), thereby defining the modelled fROMs. Indeed, the model is capable of computing its own inverse kinematics (IK) by taking into account the ICs. Besides, to estimate the configuration of the upper limb, %compute the pose of the human hand, 
given an impairment and a goal pose, we run an underlying optimisation process from \cite{lagomarsino2025mitigating}, within which the ICs are incorporated, as well. In Fig. \ref{pipeline}, this computation is represented by the \textit{IK Human Predictor} block.  
In particular, the optimisation solves
\begin{equation}
\begin{aligned}
    \min_{\boldsymbol{q}} \,
    & \left\| \bm{A}(\boldsymbol{q}-\boldsymbol{q}_{m}) \right\|^2
    + \gamma \left\|(\mathbf{I}-\bm{A})(\boldsymbol{q}-\boldsymbol{q}_{n})\right\|^2,
    \\
    & \text{s.t. } (\ref{eq1}), \boldsymbol{x}_{obj}(\boldsymbol{q})=\boldsymbol{h},
    %\label{eq:optimisation}
\end{aligned}
\end{equation}

where the first term minimises the deviation of fully or partially impaired joints from their current configuration. This is achieved through the diagonal weighting matrix $\bm{A}=\mathrm{diag}(\bm{\alpha})$, whose entries correspond to the ICs and assign larger penalties to the motion of more severely impaired joints. The second term instead penalises compensatory movements by minimising the deviation of the healthy joints from the nominal configuration $\bm{q}_n$. The scalar parameter $\gamma\in[0,1]$ weights this contribution, giving priority to preserving the current configuration of impaired joints while softly encouraging comfortable postures for the unconstrained joints. In our implementation, $\gamma$ was set to 0.05.

The task-specific constraints enforce that the object pose $\bm{x}_{obj}$, which corresponds to the grasp pose reached by the user's hand during the interaction, matches the desired task pose. Within the proposed active perception framework, the desired task pose $\boldsymbol{h}$ is iteratively updated to explore different handover configurations, thereby eliciting informative upper-limb motions that improve the estimation of the user's ICs.

This model was chosen under the strong assumption that only the movements of the upper limb were relevant during an activity to determine the ROMs of its joints. In real-life settings, the active perception process needs to be determined outside the regular working activity, to enable reproducible conditions for the active perception process and avoid the interference of compensatory movements \cite{lagomarsino2025mitigating, Scano2025}. As a consequence, it is assumed that during the active perception process the user is told to hold its position in space and to use only the movements of their arm for the reaching movements \cite{rocha2025post}. This prevents the measurements from being affected by displacement of the person and trunk compensation \cite{lagomarsino2026towards,Scano2025}.

\begin{comment}
    
\begin{table}[H]
\centering
\caption{Denavit-Hartenberg parameters of the adopted human upper-limb kinematic model (see also Fig. \ref{kinematic_chain}). Only in this case, $\alpha_i$ does not refer to the ICs.}
\label{tab:dh-parameters}
\begin{tabular}{|c |c |c |c |c|}
\hline
$\bm{i}$ & $\bm{\theta_i}$ & $\bm{a_i}$ & $\bm{$\alpha_i$} & \bm{$d_i$} \\
\hline
1 & $q_1$                & $l_{\mathrm{spine}}^{z}$      & $-\frac{\pi}{2}$ & $-l_{\mathrm{spine}}^{y}$ \\
\hline
2 & $q_2$                & $0$                            & $\frac{\pi}{2}$  & $-l_{\mathrm{spine}}^{x}$ \\
\hline
3 & $q_3+\pi$           & $0$                            & $\frac{\pi}{2}$  & $0$ \\
\hline
4 & $q_4-\frac{\pi}{2}$ & $-l_{\mathrm{humerus}}^{x}$   & $\frac{\pi}{2}$  & $-l_{\mathrm{humerus}}^{z}$ \\
\hline
5 & $q_5+\pi$           & $0$                            & $\frac{\pi}{2}$  & $-l_{\mathrm{humerus}}^{y}$ \\
\hline
6 & $q_6+\frac{\pi}{2}$ & $-l_{\mathrm{radius}}^{y}$    & $\frac{\pi}{2}$  & $-l_{\mathrm{radius}}^{z}$ \\
\hline
7 & $q_7+\frac{\pi}{2}$ & $0$                            & $\frac{\pi}{2}$  & $-l_{\mathrm{radius}}^{x}$ \\
\hline
hand & $q_8$             & $0$                            & $0$               & $0$ \\
\hline
\end{tabular}%
% Replace this box with: \includegraphics[width=\linewidth]{your-diagram-screenshot}
\end{table}

\end{comment}

\subsection{Active perception}
The process iteratively estimates $\hat{\bm{\alpha}}$ via Bayesian updates through a particle filter \cite{djuric2003particle}, computed in the \textit{Belief Updater} block (see Fig. \ref{pipeline}). This choice for the core estimation algorithm was made because $\bm{\alpha}$ does not vary over time during the interaction, is characterised by non-linear relationships with the user's joint values, due to the dependence on a kinematic chain, and its components' value lie within a well-determined range (see Subsection \ref{2::mm::problem}).
The particle filter represents the posterior belief over the
ICs as the weighted empirical distribution 
\begin{equation}
p(\hat{\bm{\alpha}}\mid\mathcal{D}_t)
\approx
\sum_{i=1}^{N_p}
w_{i,t}\,
\delta(\hat{\bm{\alpha}}-\bm{\alpha}_{i,t}),
\end{equation}
where $\mathcal{D}_t$ denotes the observations collected up to
iteration $t$. For reporting and termination purposes, this belief
is summarised through its weighted mean
$\bm{\mu}_{\bm{\alpha},t}$ and marginal variance vector
$\bm{\sigma}_{\bm{\alpha},t}^{2}$ . 

%The process iteratively estimates $\hat{\bm{\alpha}}$ as a $\mathcal{N}\!\left(
%\bm{\mu}_{\bm{\alpha}},
%\operatorname{diag}(\bm{\sigma}_{\bm{\alpha}}^2)
%\right)$ via Bayesian updates through a particle filter \cite{djuric2003particle}, computed in the \textit{Belief Updater} block (see Fig. \ref{pipeline}). This choice for the core estimation algorithm was made because $\bm{\alpha}$ does not vary over time during the interaction, is characterised by non-linear relationships with the user's joint values, due to the dependence on a kinematic chain, and its components' value lie within a well-determined range (see Subsection \ref{2::mm::problem}).

The robot starts the active perception requesting a handover to the user by moving the object to a handover pose $\bm{h}_t$. When the user finishes reaching out to the object, the system detects its kinematics of the user's upper limb $\bm{q}_t$ and uses it to define a log-likelihood distribution of the IK of the user at $\bm{h}_t$ as 

\begin{equation}
\begin{aligned}
\mathcal{LL}(\bm{q}_t) = ln(\mathcal{N}(\bm{q}_{N_o,t},\sigma_q^2 \bm{I}_{N_o})), N_o \leq N_j,
\end{aligned}
\end{equation}
with $\bm{q}_{N_o,t}$ a vector of the components of $\bm{q}_t$ used to define the log-likelihood upon and $\bm{I}_{N_o}$ as $N_o \times N_o$ identity matrix. In our implementation, we set $\sigma_q^2$ to 0.2 and considered the same DOFs related to $\hat{\bm{\alpha}}$ (see Subsection \ref{2::mm::problem}). At the same time, a set of $N_p = 50$ different particles $\mathcal{A}_t = \{\bm{\alpha}_{i,t}\}_{i=1}^{N_p}$, that is random instances of the IC vector to estimate, are defined and assigned each an equal weight of $1/N_p$, collected in a vector $\bm{W}_t$. In the Bayesian framework, the particle set $\mathcal{A}_t$ together with the associated weight vector $\bm{W}_t$ represents the prior distribution of $\bm{\alpha}$ at time $t$. The pipeline then computes the IK solutions of the human model for pose $\bm{h}_t$ given $\mathcal{A}_t$ (see Subsection \ref{2::mm::human-model}), defined as  

\begin{equation}
\begin{aligned}
\mathcal{Q}_t
=
\left\{
IK(\bm{h}_t \mid \bm{\alpha}_{i,t})
\right\}_{i=1}^{N_p}.
\end{aligned}
\end{equation}

This produces a set of candidates IK solutions for the internal human model, given the impairment belief based on the particles and their weights at time $t$. For these joint vectors, the log-probability of each of them is evaluated and used to compute the quantity

\begin{equation}
\begin{aligned}
\ln(\bm{W}_{t+1})
&=
\bm{L}_t
+
\ln(\bm{W}_t+\epsilon),
\\[2mm]
\bm{L}_t
&=
\left[
\mathcal{LL}(\bm{q}_i)
\right]_{\bm{q}_i\in\mathcal{Q}_t},
\end{aligned}
\end{equation}

%\begin{equation}
%\begin{aligned}
%\bm{W}_t = [P(\bm{q}<\bm{q}_i) = \int_{-\infty}^{\bm{q}_i} \mathcal{L}(\bm{q})\,d\bm{q}], \forall \bm{q}_i \in \bm{Q}_t
%\end{aligned}
%\end{equation}

%\begin{equation}
%\begin{aligned}
%\bm{W}_t = [\mathcal{LL}(\bm{q}_i)], \forall \bm{q}_i \in \bm{Q}_t
%\end{aligned}
%\end{equation}

where \(0<\epsilon\ll1\) is a small positive constant introduced to ensure numerical stability when evaluating the logarithm. Summing the log-likelihood  terms with the logarithm of the previous weights allows them to be weighted on the weights the particles had prior to the update. Upon normalisation of $ln(\bm{W}_{t+1})$ in the log space and exponentiation, the vector $\bm{W}_{t+1}$ is obtained and becomes the new weight vector associated with the particles. At this point, the belief on $\bm{\hat{\bm{\alpha}}}$ is updated as follows:

\begin{equation}
\begin{aligned}
\bm{\mu}_{\bm{\alpha},t} 
&= 
\frac{\sum_{i=1}^{N_p} w_{i,t+1} \bm{\alpha}_{i,t}}{\sum_{i=1}^{N_p} w_{i,t+1}},  
\\[2mm]
\bm{\sigma}_{\bm{\alpha},t}^2 
&= \frac{\sum_{i=1}^{N_p} w_{i,t+1} (\bm{\alpha}_{i,t} - \bm{\mu}_{\bm{\alpha},t})^2}{\sum_{i=1}^{N_p} w_{i,t+1}},
\end{aligned}
\end{equation}
where $w_{i,t+1}$ are the components of $\bm{W}_{t+1}$. After having performed the update, we check whether $\bm{\sigma}_{\bm{\alpha},t}^2 < \sigma_{thr}^2$ to guarantee the precision of the estimate. In our design, we set $\sigma_{thr}^2 = 0.003$, so that the error would not be bigger than 0.1 for the majority of the times. If this condition is met for three consecutive iterations of the process just described, the estimate of $\hat{\bm{\alpha}} = \bm{\mu}_{\bm{\alpha},t}$ is accepted.  Otherwise, we generate a new handover request to the user. If the termination condition is not met after 25 handovers, the current estimate $\hat{\bm{\alpha}}=\bm{\mu}_{\bm{\alpha},t}$ is accepted anyway. This upper limit was introduced to keep the duration of the estimation procedure acceptable for a human--robot interaction. The estimation process is updated after each handover rather than at a fixed frequency, with the overall interaction duration depending primarily on the robot's movements and the time required for the user to reach the object. In our implementation, robot trajectories between consecutive handover poses were assigned a duration of 5~s. Although the computational time of the IK predictions and particle filter updates was not measured independently, these computations were performed online as part of the handover interaction.

During the estimation, some of the particles could be assigned weights close to 0, thus not being so significant for the estimation. When many particles do not influence the estimate anymore, the elements of $\mathcal{A}_t$ all undergo a resampling process. For this purpose, we define the quantity

\begin{equation}
\begin{aligned}
N_{\mathrm{eff}_t}
&= 
\frac{1}{\sum_{i=1}^{N_p} w_{i,t}^2}  
\end{aligned}
\end{equation}
as indicator of the amount of particles that have non-negligible weight, thus effective to the estimation. When the quantity $N_{\mathrm{eff},t}/N_p$ is under a certain threshold (0.4 in our implementation), all the particles are resampled with replacement weighted on $\bm{W}_t$, perturbed with Gaussian noise $\mathcal{N}\!\left(
\bm{0},
\sigma_{jitt}^2\bm{I}_{N_{imp}}
\right)$ and assigned $1/N_p$ weights each again. In our implementation, $\sigma_{jitt}^2=0.05$.   
\label{2::mm::active-perception}

\begin{figure*}[!t]
      \vspace{5mm}
      \centering
      \includegraphics[scale=0.56]{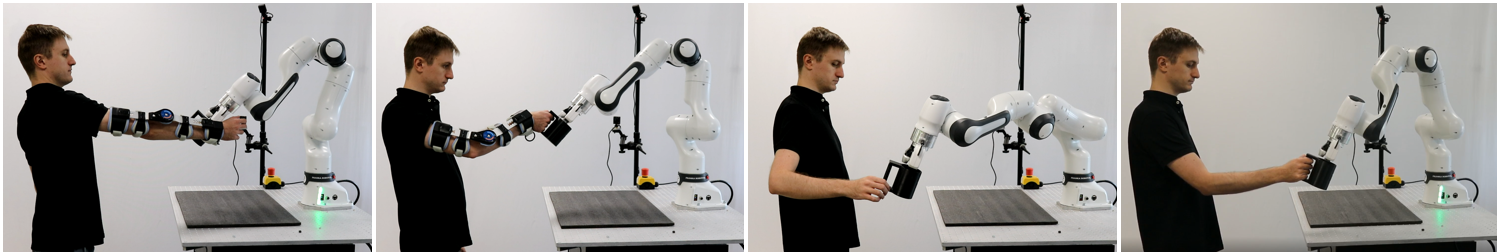}
      \caption{Pictures from the physical validations with enacted impairments. From left to right: $\alpha_{S_y}^*=1.0$, $\alpha_{S_y}^*=0.5$, $\alpha_{E_y}^*=1.0$ and $\alpha_{E_y}^*=0.5$. In some instances of the interactions, the participant was not able to reach the object. This is expected, as the pipeline does not have prior knowledge of the user's real impairment. Nonetheless, the pipeline is able to perform the updates, because the underlying optimisation used enforces the constraint $\bm{x}_{obj}(\bm{q})=\bm{h}_t$ at the best extent allowed by the impairment, but it is able to return an IK solution even when the constraint cannot be met (see Subsection \ref{2::mm::human-model}).}
      \label{snapshots}
   \end{figure*}
\section{EXPERIMENTS}
The active perception pipeline was evaluated in simulated and
physical experiments by estimating the ICs associated with shoulder
and elbow flexion \cite{rocha2025post} (referring to Fig. \ref{kinematic_chain}, $i=2$ and $i=4$; from this point on, we refer to them as $\alpha_{S_y}$ and $\alpha_{E_y}$, respectively). The ICs of all remaining upper-limb DOFs were fixed to 0, corresponding to unrestricted healthy ROMs in the model \cite{rocha2025post}. In simulation, all possible combinations of their related ICs, ranging from 0 to 1 with a step of 0.025641, were estimated, for a total of 1600 2D vectors. In this case, the recordings in real-life were replaced by another instance of the \textit{IK Human Predictor} block (see Subsection \ref{2::mm::human-model})    \begin{figure}[htbp!]
      \centering
      \includegraphics[scale=0.25]{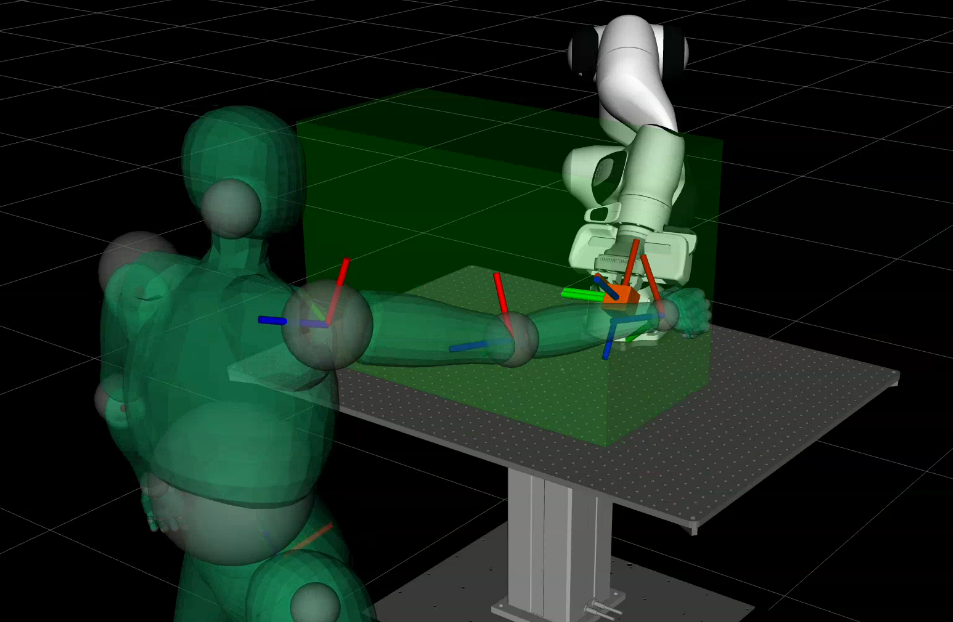}
      \caption{Rendering of the experimental setting in Rviz 2. The human model was being acquired through the Real-Move setup. The green cube represents the action space the robot ould requests handovers within. The orange cube centered in the frame with the blue z axis oriented towards the user's head represents $\bm{h}_t$ and the mug handle used in the handovers.}
      \label{rviz}
   \end{figure} running uniquely with the configuration of ICs to estimate at that time for $\alpha_{S_y}$ and $\alpha_{E_y}$ and with the ICs of the remaining DOFs fixed to 0. 
%The active perception pipeline was evaluated through simulated and real-life experiments, in which we estimated the mobility of the shoulder and elbow flexion (referring to Fig. \ref{kinematic_chain}, $i=2$ and $i=4$; from this point on, we refer to them as $\alpha_{S_y}$ and $\alpha_{E_y}$, respectively), assuming full mobility for all the other arm joints \cite{rocha2025post}. In this setting, the processed $\bm{q}_t$ was composed by the joint values of the related DOFs to estimate. In simulation, all possible combinations of their related ICs, ranging from 0 to 1 with a step of 0.025641, were estimated, for a total of 1600 2D vectors. In this case, the recordings in real-life were replaced by another instance of the \textit{IK Human Predictor} block (see Subsection \ref{2::mm::human-model}) running uniquely with the configuration of ICs to estimate at that time for elbow and shoulder flexion and the other ROMs set as described. 

The real-life experiments were performed with a participant (male, age: 32, height: 1.80 m) who, additionally to the case of moving the arm freely while standing still, enacted half and full impairment of either $\alpha_{S_y}$ or $\alpha_{E_y}$ (see Fig. \ref{snapshots}). To reliably reproduce the impairment on $\alpha_{E_y}$, they wore a harness that could bound it within different tunable ranges \cite{Liu2025, grant2021use} (see Figg. \ref{figurelabel} and \ref{snapshots}). In the experimental setting, a Franka Robotics® (Munich, Germany) Franka Research 3 was used for performing the handovers during the active perception process. At the startup, the space $\bm{h}_t$ could be selected from a box space whose position in the workspace was tailored on the height and reachability of the user (see Fig. \ref{rviz}) \cite{ISO14738_2002, semeraro2025trihrcbot}. The handover poses were chosen randomly within the set space. The robot initially grabbed a mug and then moved it so that a reference frame on the handle coincided with the selected handover pose (see Fig. \ref{rviz}). The joint values and poses of the participants were detected through Real-Move (Real-Move\textsuperscript{TM}, Genoa, Italy), markerless motion capture system \cite{real-move}. Their poses were used at the startup to tailor the action space, while their values were processed only when the participant reached the handover pose to the closest possible extent permitted by the enacted impairment. More precisely, the skeleton upper-limb joint values resulting from the circular average of the 7 recordings following the experimenter's recording trigger corresponded to the $\bm{q}_t$ vector processed by the pipeline. The skeleton detection performed by Real-Move was compatible with the rendering of our human model running in the \textit{IK Human Predictor} block (see Subsection \ref{2::mm::human-model}), which was scaled to the participant's bodily features prior to beginning (see Subsection \ref{2::mm::human-model}). While the estimate was being updated, the participant reconvened every time to their starting resting stance, waiting for the next handover request.

The whole ensemble was controlled through a ROS 2 Humble workspace running on Ubuntu 22.04 (12\textsuperscript{th} Gen Intel® Core™ i7-12700H × 20). The \textit{IK Human Predictor} block (see Subsection \ref{2::mm::human-model}) ran on ROS Noetic and communicated with the remaining system through a ROS bridge. The experiments were carried out at the HRI$^2$ Laboratory, in accordance with the Helsinki Declaration, and the protocol was approved by the Comitato etico territoriale Regione Liguria (Protocol IIT\_HRII\_002).

%Experiments were carried out at the HRII Lab., in accordance with the Helsinki Declaration, and the protocol was approved by the Comitato etico territoriale Regione Liguria (Protocol IIT\_HRII\_002).

\label{3::exp}
\section{RESULTS}
The experiments in simulation were used to report statistics on the performance of the active perception pipeline, as they rely on the possibility of reiterating the experiment numerous times and under many impairment settings. Regarding the real-life recordings, instead, they relate to fewer instances, so they were used to show the progression of the active perception process over time. The results are shown below.
\label{4::results}

\subsection{Simulation}
\label{4::results::simulation}

Table \ref{performance_metrics} reports the performance of the IC estimation process (see Section \ref{3::exp}). The results were grouped according to low and high ground truth ICs, with values up to and including 0.5 classified as low impairment and values above 0.5 as high impairment. The error was computed as the mean of the components of $|\bm{\hat{\bm{\alpha}}}-\bm{\alpha^*}|$, where $\bm{\alpha^*}$ denotes the ground truth vector of the ICs to estimate.

\begin{table}[htbp!]
\caption{Performance metrics across shoulder (S) and elbow (E) flexion impairment configurations. Each configuration was evaluated over 400 trials.}
\label{performance_metrics}
\centering
\renewcommand{\arraystretch}{1.25}
\setlength{\tabcolsep}{3pt}

\begin{minipage}{\columnwidth}
\centering
\begin{tabular}{|C{0.34\columnwidth}|
                C{0.1\columnwidth}|C{0.1\columnwidth}|
                C{0.09\columnwidth}|C{0.09\columnwidth}|}
\hline
\multirow{2}{=}{\centering\textbf{Configuration}} &
\multicolumn{2}{c|}{\textbf{Error}} &
\multicolumn{2}{c|}{\textbf{N. of Handovers}} \\
\cline{2-5}
& \textbf{Mean} & \multicolumn{1}{c|}{\textbf{SD}}
& \textbf{Mean} & \multicolumn{1}{c|}{\textbf{SD}} \\
\hline
$\alpha_{S_y}^*\leq0.5$, $\alpha_{E_y}^*\leq0.5$
& 0.0867 & \multicolumn{1}{c|}{0.0867}
& 23.86 & \multicolumn{1}{c|}{3.51} \\
\hline
$\alpha_{S_y}^*\leq0.5$, $\alpha_{E_y}^*>0.5$
& 0.0504 & \multicolumn{1}{c|}{0.0582}
& 21.56 & \multicolumn{1}{c|}{5.84} \\
\hline
$\alpha_{S_y}^*>0.5$, $\alpha_{E_y}^*\leq0.5$
& 0.0833 & \multicolumn{1}{c|}{0.0615}
& 23.62 & \multicolumn{1}{c|}{4.11} \\
\hline
$\alpha_{S_y}^*>0.5$, $\alpha_{E_y}^*>0.5$
& 0.0453 & \multicolumn{1}{c|}{0.0590}
& 19.91 & \multicolumn{1}{c|}{6.80} \\
\hline
\end{tabular}
\end{minipage}

\vspace{2mm}

\begin{minipage}{\columnwidth}
\centering
\begin{tabular}{|C{0.34\columnwidth}|
                C{0.1\columnwidth}|C{0.1\columnwidth}|
                C{0.23\columnwidth}|}
\hline
\multirow{2}{=}{\centering\textbf{Configuration}} &
\multicolumn{2}{c|}{\textbf{N. of Resamplings}} &
\multirow{2}{=}{\centering\textbf{N. of Early Terminations}} \\
\cline{2-3}
& \textbf{Mean} & \multicolumn{1}{c|}{\textbf{SD}} & \\
\hline
$\alpha_{S_y}^*\leq0.5$, $\alpha_{E_y}^*\leq0.5$
& 3.19 & \multicolumn{1}{c|}{1.32} & 57 \\
\hline
$\alpha_{S_y}^*\leq0.5$, $\alpha_{E_y}^*>0.5$
& 5.25 & \multicolumn{1}{c|}{2.15} & 136 \\
\hline
$\alpha_{S_y}^*>0.5$, $\alpha_{E_y}^*\leq0.5$
& 3.27 & \multicolumn{1}{c|}{1.32} & 54 \\
\hline
$\alpha_{S_y}^*>0.5$, $\alpha_{E_y}^*>0.5$
& 5.07 & \multicolumn{1}{c|}{2.37} & 187 \\
\hline
\end{tabular}
\end{minipage}
\end{table}

Estimation performance was lower for configurations with low $\alpha_{E_y}^*$, which exhibited higher mean errors regardless of $\alpha_{S_y}^*$. These configurations also required more handovers on average and resulted in fewer early terminations. In particular, the mean number of handovers exceeded 23 for both configurations with low $\alpha_{E_y}^*$, compared with approximately 21 when $\alpha_{E_y}^*$ was high. Similarly, early termination occurred on average only in 13.75\% of the related trials for low $\alpha_{E_y}^*$, compared with 40.25\% of the trials with high $\alpha_{E_y}^*$. The mean number of resamplings was also lower in the former configurations, remaining close to 3.2 compared with approximately 5.2 in the latter. Overall, these results indicate that lower values of $\alpha_{E_y}^*$ were associated with more frequent failure to meet the termination condition within 25 handovers, while variations in $\alpha_{S_y}^*$ had a smaller influence.

\subsection{Physical validation}
\label{4::results::validation}

\begin{figure}[htbp!]
 %     \centering
      \hspace{-5mm}
      
      \includegraphics[scale=0.4]{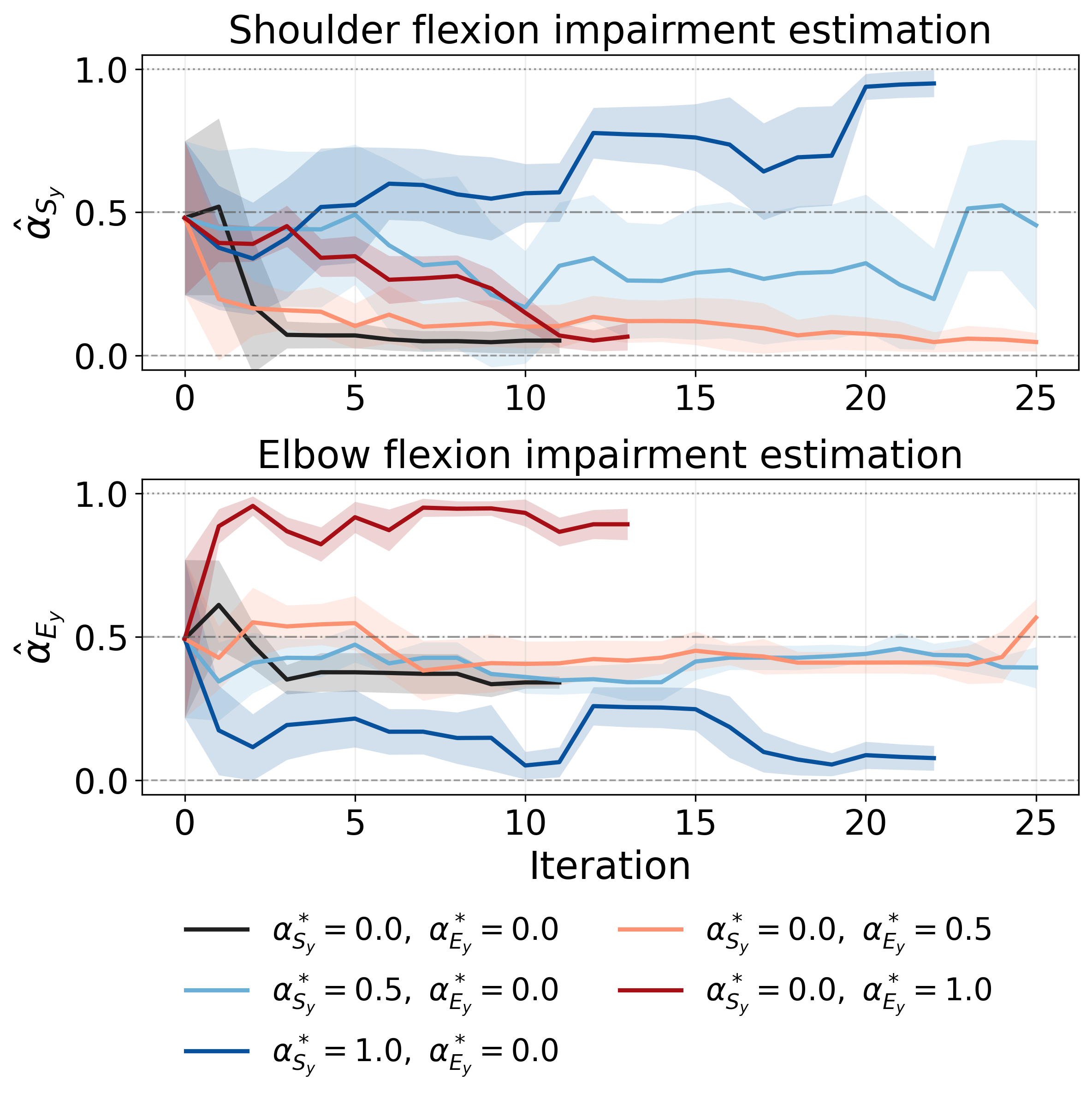}
      \caption{Trends on the joint estimate of $\alpha_{S_y}^*$ and $\alpha_{E_y}^*$ in the physical validations. They all start from 0.5 because of the way $\mathcal{A}_t$ was defined (see Subsection \ref{2::mm::active-perception}). The shaded areas were plotted based on the related $\bm{\sigma_{\bm{\alpha},t}}$. Trends terminating before iteration 25 indicate that the run was able to meet the early termination criterion (see Subsection \ref{2::mm::active-perception}).}
      \label{online-validations}
   \end{figure}

Fig. \ref{online-validations} shows the evolution of the estimated ICs over the course of the interactions, for the impairment configurations explored (see Section \ref{3::exp}). For three of the recordings, the pipeline was able to converge to a precise estimate before reaching the highest number of iterations permitted. \begin{figure}[htbp!]
      \centering
      %\hspace{-7}
      \includegraphics[scale=0.41]{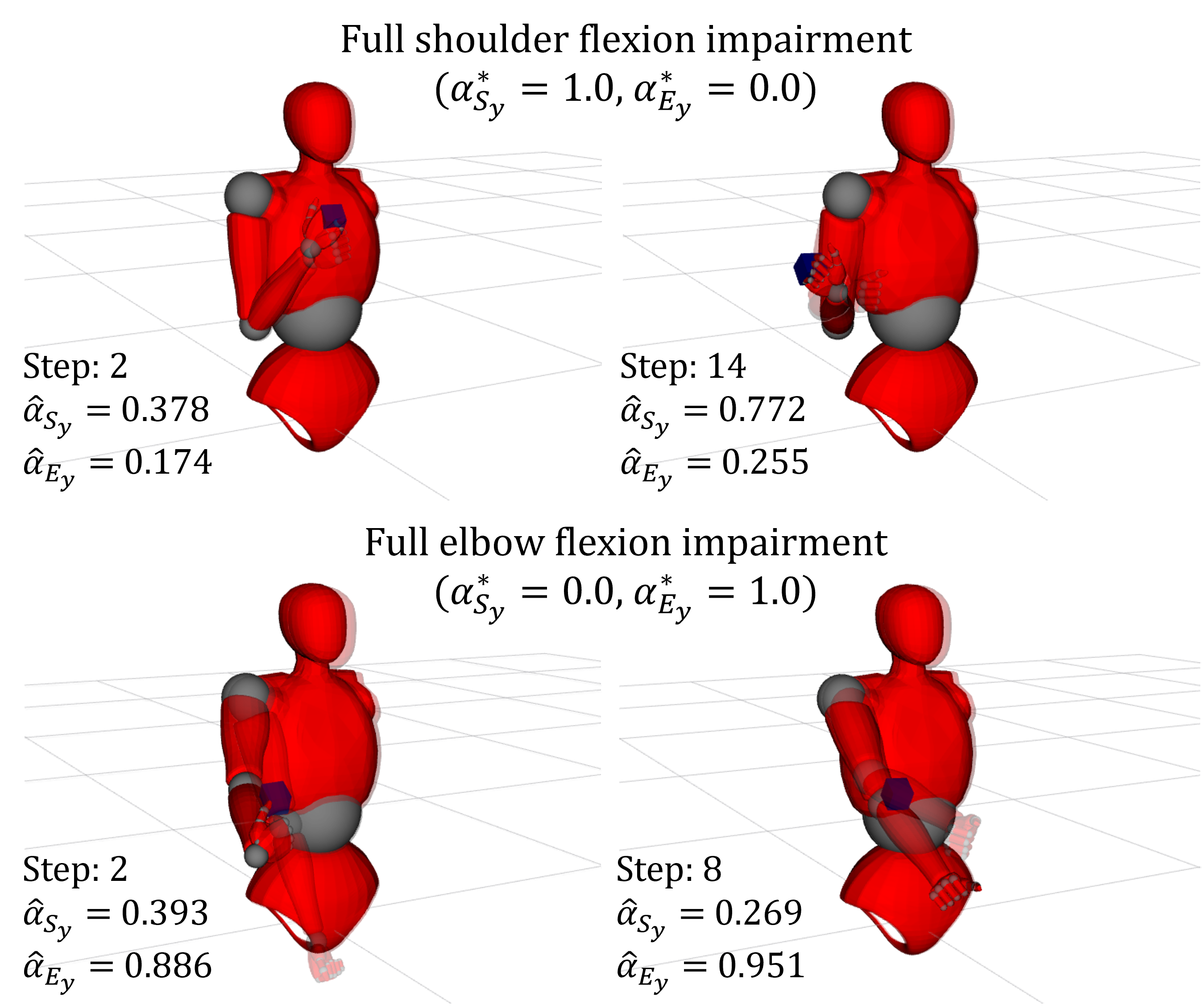}
      \caption{Visual comparisons between the result of the IK solution given by the $\bm{\hat{\bm{\alpha}}}$ at the current iteration and the recorded joint values (shaded mannequin) of the position performed by the participant during the handover. %(the phantom).
      }
      \label{phantoms}
   \end{figure} On the whole, the pipeline was able to understand the magnitude relationship between the two impairments being estimated from the first iterations. Fig. \ref{phantoms} qualitatively illustrates two iterations of the proposed pipeline, highlighting the agreement between the measured postures and the reconstructed ones using the current ICs estimate, which drives the subsequent estimate update. For $\alpha_{E_y}^*=1.0$, the pipeline correctly identified the presence of an elbow flexion limitation as early as the second iteration. However, the reconstructed posture still exhibited a noticeable discrepancy from the measured one. By iteration 8, the estimated posture closely matched the actual human configuration, resulting in early convergence at iteration 11 (see Fig. \ref{online-validations}). 
Conversely, for $\alpha_{S_y}^*=1.0$, the algorithm initially underestimated the severity of the limitation. This behaviour can be explained by the fact that the handover pose selected during the early iterations primarily required elbow flexion (see Fig. \ref{phantoms}), providing limited information about the subject's shoulder mobility. As a result, additional observations were needed before accurately estimating $\alpha_{S_y}$, leading to convergence after 22 iterations (see Fig. \ref{online-validations}). On the other hand, the estimate appears less accurate in the cases of $\alpha_{E_y}^* = 0$ and low $\alpha_{S_y}^*$. Specifically, $\hat{\alpha}_{E_y}$ is off by 0.4 with respect to $\alpha_{E_y}^*$, while $\hat{\alpha}_{S_y}$ is much more accurate at the process termination, even with different values of $\alpha_{S_y}^*$.

\section{DISCUSSION}
\label{5::discussion}

The results demonstrate both the potential of the proposed active
perception pipeline and the main limitations that currently affect its estimation performance. They are discussed below, followed by broader considerations.

\subsection{Estimation performance and limitations}
\label{5::discussions::performance}

This work presented a pipeline for actively estimating the
impairment coefficients that parameterise a user's task-available
upper-limb fROM through repeated human-robot handovers. With a simulated human model with the same underlying IK optimisation (see Subsection \ref{2::mm::human-model}), this active perception process was capable of estimating the ICs associated with shoulder and elbow flexion with an error on average up to a tenth of the range of possible values. This was achieved on average in less than 25 handover requests, thanks to the early termination condition (see Subsection \ref{2::mm::active-perception}). These results were achieved without any learning previously performed, by only relying on the online estimate of the ICs through the kinematics of the user at the handover poses. All these results make the pipeline easily deployable in a robot to update its internal representation of the user's functional
joint limits when needed, for a more personalised collaboration (see Section \ref{1::intro}). 
%This work presented a pipeline for actively estimating a user's upper-limb ROMs through repeated human-robot handovers. With a simulated human model with the same underlying IK optimisation (see Subsection \ref{2::mm::human-model}), this active perception process is capable of estimating the impairments of shoulder and elbow flexion with an error on average lower than a tenth of the range of possible values. This was achieved on average with less than 25 handovers requests, thanks to the early termination condition (see Subsection \ref{2::mm::active-perception}). These results were achieved without any learning previously performed, by only relying on the online estimate of the ICs through the kinematics of the user at the handover points. All these results make the pipeline easily deployable in a robot to update the mobility of the user when needed, for a more personalised collaboration (see Section \ref{1::intro}). 

Despite these promising features, several opportunities remain for improving the proposed approach. Looking at the online validations (see Subsection \ref{4::results::validation}), the pipeline was less accurate when $\alpha_{E_y}^*$ was set to 0 and $\alpha_{S_y}^*$ was low. This can be explained by the following consideration. To understand that the elbow is not impaired, the pipeline would need to elicit configurations spanning a sufficiently large portion of the elbow's available fROM to perform the right estimate under this setting. As the shoulder is a more proximal joint in the arm kinematic chain, its only movement already allows a person to greatly reduce the distance from their hand to the target point. When this happens, less movement is required by the elbow to reach the target. This means that the pipeline did not manage to elicit enough elbow excursion through the handovers to constrain the corresponding fROM accurately when the shoulder was free to move. This observation is confirmed by the results achieved in simulation (see Subsection \ref{4::results::simulation}), that show similar performance decrease with low $\alpha_{E_y}^*$. 

\subsection{Future developments}

\label{5::discussions::developments}

The limitation previously reported (see Subsection \ref{5::discussions::performance}) was mainly due by the choice of the action space (see Section \ref{3::exp}), and could be
addressed by modifying its geometry so as to elicit upper-limb
configurations requiring a larger contribution from elbow flexion
\cite{mehler2017minimizing}. The action space adopted in this study
(see Fig.~\ref{rviz}) may therefore be too limited, and wider or
more complex spatial regions may be needed. Any such redesign would
also have to account for the reachable workspace of the robotic
platform, since not all informative target poses may be physically
accessible.

Secondly, the handover poses used during the estimation were randomly selected from the designated action space (see Section~\ref{3::exp}). Nevertheless, the proposed framework constitutes an active perception approach in the sense that the robot deliberately elicits reaching movements by presenting target poses to the user, rather than relying solely on observations of independently performed activities. By controlling the target pose, the robot establishes a known reference against which the measured joint configuration can be compared with the predictions of its internal human model, allowing the belief over the ICs to be updated. Thus, although the current implementation does not explicitly optimise the informativeness of individual handovers, the acquisition of observations remains robot-initiated.

Future work should investigate an information-driven policy that selects the next handover pose according to the current belief over the user's ICs. Such a policy could maximize the expected information gained from each interaction and reduce the number of handovers required to obtain an accurate fROM estimate. Secondly, the experimental validation focused on shoulder and elbow flexion, although the human model includes ICs for additional upper-limb DOFs (see Subsection~\ref{2::mm::human-model}). Extending the estimation framework to infer a broader set of ICs would provide the robot with a richer representation of the user's functional capabilities, enabling more personalized human-robot collaboration \cite{lagomarsino2026towards}. Finally, the current upper-limb model considers a kinematic chain with 7 DOFs (see Subsection \ref{2::mm::human-model}). Extending the model to include additional DOFs could increase the information extracted from each handover and improve the fidelity of the estimated mobility profile. However, this extension would also require reliable sensing of the additional joint variables included in the model.

\subsection{Deployment on humanoids}
\label{5::discussions::humanoids}
Beyond the specific handover implementation, a broader conceptual contribution of this work lies in framing fROM estimation as an active perception problem based on guided reaching. The proposed formulation is not tied to a specific robot platform: it requires an embodied system capable of presenting target poses that elicit informative upper-limb configurations and observing the resulting human motion. From the interaction perspective, the robot acts as a spatially situated target presenter, while its own kinematic constraints determine which handover poses can be physically realised. Consequently, if a desired pose is unreachable, the active perception process can remain applicable by selecting an alternative reachable pose with comparable information value.

This formulation is particularly relevant for humanoid robots and mobile manipulators, whose larger and more flexible reachable workspaces may allow them to present targets around the user that cannot be covered by a fixed-base manipulator. In principle, the same guided reaching abstraction could also be extended to complementary interfaces, such as augmented reality systems that visually indicate target locations \cite{letourneau2024augmented,basiratzadeh2020augmented}.
However, embodied robotic platforms retain important advantages for workplace deployment, including the ability to physically manipulate task objects, adapt the interaction geometry online, and integrate the estimated mobility profile directly into subsequent collaborative actions.

\addtolength{\textheight}{-0.5cm}   % This command serves to balance the column lengths
                                  % on the last page of the document manually. It shortens
                                  % the textheight of the last page by a suitable amount.
                                  % This command does not take effect until the next page
                                  % so it should come on the page before the last. Make
                                  % sure that you do not shorten the textheight too much. 

%%%%%%%%%%%%%%%%%%%%%%%%%%%%%%%%%%%%%%%%%%%%%%%%%%%%%%%%%%%%%%%%%%%%%%%%%%%%%%%%

%%%%%%%%%%%%%%%%%%%%%%%%%%%%%%%%%%%%%%%%%%%%%%%%%%%%%%%%%%%%%%%%%%%%%%%%%%%%%%%%

\section{CONCLUSIONS}

This work introduced a method for a robot to autonomously estimate, through handovers, the impairment coefficients that parameterise a user's goal-oriented functional ranges of motion of the upper limb. These estimates could subsequently be used to adapt human--robot interactions to the user's physical capabilities and support an active recovery process. Specifically, an active perception pipeline that estimates the impairment coefficients through a particle filter was described. The process was tested in simulation through a human model whose functional joint limits were parameterised by the same representation based on impairment coefficients. In real-life settings, the pipeline was validated by having a participant perform repetitive handovers with a robotic manipulator while wearing harnesses to partially prevent their movements within delimited ranges. The robot jointly estimated the impairment coefficients of the shoulder and elbow within 25 handovers.

Although a robotic manipulator was used in this study, the approach is especially relevant to humanoid robots, whose human-like morphology, whole-body capabilities, and potential for natural interaction make them particularly suitable for return-to-work scenarios. Unlike a fixed manipulator, a humanoid robot could adapt not only the handover trajectory but also its posture, positioning, and interaction strategy to the user’s functional limitations, enabling more personalised, socially engaging, and ecologically valid rehabilitation activities.
\label{6::conclusions}

\section*{ACKNOWLEDGMENTS}
The authors acknowledge Mattia Leonori and Luca Fortini for the assistance with the Real-Move setup, and Luigi Pedata for the assistance with the robotic platform.

%%%%%%%%%%%%%%%%%%%%%%%%%%%%%%%%%%%%%%%%%%%%%%%%%%%%%%%%%%%%%%%%%%%%%%%%%%%%%%%%

\bibliographystyle{IEEEtran}
\bibliography{IEEE_example}

\end{document}